\RequirePackage[hyphens]{url}

\documentclass[11pt,a4paper]{article}
\usepackage{algorithm}
\usepackage[noend]{algpseudocode}
\usepackage{algorithmicx}
\usepackage{amssymb}
\usepackage{caption}
\usepackage{color}
\usepackage{CJKutf8}
\usepackage[hyperref]{emnlp-ijcnlp-2019}
\usepackage{graphicx}
\usepackage{latexsym}
\usepackage{listings}
\usepackage{multirow}
\usepackage{soul}
\usepackage{times}
\usepackage{xcolor}
\usepackage{booktabs}
\usepackage{hyperref}
\usepackage{flafter}
\usepackage[bottom]{footmisc}

\newcommand*\Let[2]{\State #1 $\gets$ #2}
\algrenewcommand\algorithmicrequire{\textbf{Input:}}
\algrenewcommand\algorithmicensure{\textbf{Output:}}

\definecolor{codegreen}{rgb}{0,0.6,0}
\definecolor{codegray}{rgb}{0.5,0.5,0.5}
\definecolor{codepurple}{rgb}{0.58,0,0.82}
\definecolor{paperred}{rgb}{.84,0,0,}
\definecolor{greybackground}{rgb}{.98,0.98,0.98}

\lstdefinestyle{mystyle}{
    backgroundcolor=\color{greybackground},   
    commentstyle=\color{codegreen},
    keywordstyle=\color{purple},
    numberstyle=\tiny\color{codegray},
    stringstyle=\color{paperred},
    breakatwhitespace=false,         
    breaklines=true,                 
    captionpos=b,                    
    keepspaces=true,                 
    numbersep=5pt,                  
    showspaces=false,                
    showstringspaces=false,
    showtabs=false,                  
    tabsize=2,frame=single,
    basicstyle=\scriptsize\ttfamily,
    belowskip=-1.5em,
}
\aclfinalcopy 

\title{ReQA: An Evaluation for End-to-End Answer Retrieval Models}

\author{Amin Ahmad, Noah Constant, Yinfei Yang, Daniel Cer \\
    Google Research, Mountain View, USA \\
    {\tt \{aahmad, nconstant, yinfeiy, cer\}@google.com} \\}

\date{}

\begin{document}
\maketitle


\begin{abstract}

Popular QA benchmarks like SQuAD have driven progress on the task of identifying answer spans within a specific passage, with models now surpassing human performance.
However, retrieving relevant answers from a huge corpus of documents is still a challenging problem, and places different requirements on the model architecture.
There is growing interest in developing scalable answer retrieval models trained end-to-end, bypassing the typical document retrieval step.
In this paper, we introduce Retrieval~Question-Answering~(ReQA), a benchmark for evaluating large-scale sentence-level answer retrieval models.
We establish baselines using both neural encoding models as well as classical information retrieval techniques.
We release our evaluation code to encourage further work on this challenging task.

\end{abstract}


\section{Introduction}

Popular QA benchmarks like SQuAD \citep{rajpurkar-etal-2016-squad} have driven impressive progress on the task of identifying spans of text within a specific passage that answer a posed question.
Recent models using BERT pretraining \citep{devlin-etal-2019-bert} have already surpassed human performance on SQuAD 1.1 and 2.0.

\begin{figure}[!htb]
\fbox{\parbox{0.96\columnwidth}{
\textbf{Question}: Which US county has the densest population?
\vspace{4pt} \\
\textbf{Wikipedia Page}: New York City
\vspace{4pt} \\
\textbf{Answer}: Geographically co-extensive with New York County, the borough of Manhattan's 2017 population density of 72,918 inhabitants per square mile (28,154/km\textsuperscript{2}) makes it the highest of any county in the United States and higher than the density of any individual American city.
}}
\caption{A hypothetical example of end-to-end answer retrieval, where the document containing the answer is not ``on topic'' for the question.}
\label{fig:mock}
\end{figure}

While impressive, these systems are not yet sufficient for the end task of answering user questions at scale, since in general, we don't know which documents are likely to contain an answer.
On the one hand, typical document retrieval solutions fall short here, since they aren't trained to directly model the connection between questions and answers in context.
For example, in Figure~\ref{fig:mock}, a relevant answer appears on the Wikipedia page for New York, but this document is unlikely to be retrieved, as the larger document is not highly relevant to the question.
On the other hand, QA models with strong performance on reading comprehension can't be used directly for large-scale retrieval.
This is because competitive QA models use interactions between the question and candidate answer in the early stage of modeling (e.g.~through cross-attention) making it infeasible to score a large set of candidates at inference time.

There is growing interest in training end-to-end retrieval systems that can efficiently surface relevant results without an intermediate document retrieval phase \citep{DBLP:journals/corr/abs-1811-08008,DBLP:journals/corr/abs-1810-10176,seo-etal-2019-real,henderson2019training}.
We are excited by this direction, and hope to promote further research by offering the Retrieval Question-Answering (ReQA) benchmark, which tests a model's ability to retrieve relevant answers efficiently from a large set of documents.
Our code is available at \url{https://github.com/google/retrieval-qa-eval}.

The remainder of the paper is organized as follows.
In Section~\ref{sec:objectives}, we define our goals in developing large-scale answer retrieval models.
Section~\ref{sec:method} describes our method for transforming within-document reading comprehension tasks into Retrieval Question-Answering (ReQA) tasks, and details our evaluation procedure and metrics.
Section~\ref{sec:results} describes various neural and non-neural baseline models, and characterizes their performance on several ReQA tasks.
Finally, Section~\ref{sec:related} discusses related work.


\section{Objectives}
\label{sec:objectives}

What properties would we like a large-scale answer retrieval model to have? We discuss five characteristics below that motivate the design of our evaluation.

First, we would like an \textbf{end-to-end} solution.
As illustrated in Figure \ref{fig:mock}, some answers are found in surprising places.
Pipelined systems that first retrieve topically relevant documents and then search for answer spans within only those documents risk missing good answers from documents that appear to have less overall relevance to the question.

Second, we need \textbf{efficient} retrieval, with the ability to scale to billions of answers.
Here we impose a specific condition that guarantees scalability.
We require the model to encode questions and answers \textit{independently} as high-dimensional (e.g.~512d) vectors, such that the relevance of a QA pair can be computed by taking their dot-product, as in \citet{DBLP:journals/corr/HendersonASSLGK17}.\footnote{Other distance metrics are possible. Another popular option for nearest neighbor search is cosine distance. Note, models using cosine distance can still compute relevance through a dot-product, provided the final encoding vectors are L2-normalized.}
This technique enables retrieval of relevant answers using approximate nearest neighbor search, which is sub-linear in the number of documents, and in practice close to log(N).
This condition rules out the powerful models like BERT that perform best on reading comprehension metrics.
Note, these approaches could be used to rerank a small set of retrieved candidate answers, but the evaluation of such multi-stage systems is out of the scope of this work.

Third, we focus on \textbf{sentence-level} retrieval.
In practice, sentences are a good size to present a user with a ``detailed'' answer, making it unnecessary to highlight specific spans for many use cases.\footnote{In cases where highlighting the relevant span within a sentence is important, a separate highlighting module could be learned that takes a retrieved sentence as input.}
While the experiments in this paper primarily target sentence-level retrieval, we recognize that some domains may be best served by retrieval at a different granularity, such as phrase or passage.
The evaluation techniques described in Section~\ref{sec:method} can be easily extended to cover these different granularities.

Fourth, a retrieval model should be \textbf{context aware}, in the sense that the context surrounding a sentence should affect its appropriateness as an answer. For example, an ideal QA system should be able to tell that the bolded sentence in Figure \ref{fig:context} is a good answer to the question, since the context makes it clear that ``The official language'' refers to the official language of Nigeria.

\begin{figure}[!htb]
\fbox{
\parbox{0.96\columnwidth}{
\textbf{Question}: What is Nigeria's official language? 
\vspace{4pt} \\
\textbf{Answer in Context}: [...] Nigeria has one of the largest populations of youth in the world. The country is viewed as a multinational state, as it is inhabited by over 500 ethnic groups, of which the three largest are the Hausa, Igbo and Yoruba; these ethnic groups speak over 500 different languages, and are identified with wide variety of cultures. \textbf{The official language is English.} [...]
}}
\caption{An example from SQuAD~1.1 where looking at the surrounding context is necessary to determine the relevance of the answer sentence.}
\label{fig:context}
\end{figure}

Finally, we believe a strong model should be \textbf{general purpose}, with the ability to generalize to new domains and datasets gracefully.
For this reason, \emph{we advocate using a retrieval evaluation drawn from a specific task/domain that is never used for model training.}
In the case of our tasks built on SQuAD and Natural Questions~(NQ), we evaluate on retrieval over the entire training sets, with the understanding that all data from these sets is off-limits for model training.
Additionally, we recommend not training on any Wikipedia data, as this is the source of the SQuAD and NQ document text. However, should this latter recommendation prove impractical, then, at the very least, the use of Wikipedia during training should be noted when reporting results on ReQA, being as specific as possible as to which subset was used and in what manner.
This increases our confidence that a model that evaluates well on our retrieval metrics can be applied to a wide range of open-domain QA tasks.\footnote{We strongly assert that when NLP models are used in applied systems, it is generally preferable to evaluate alternative models using data that is as distinct as reasonably possible from model training data. While this is common practice in some sub-fields of NLP such as machine translation, it is still unfortunately very common to assess other NLP models on dev and test data that is very similar to its training data (e.g., harvested from the same source using a common pipeline and a common pool of annotators). This makes it more difficult to interpret claims of models approaching ``human-level'' performance.}


\section{ReQA Evaluation}
\label{sec:method}

In this section, we describe our method for constructing \emph{Retrieval Question-Answering} (ReQA) evaluation tasks from existing machine reading based QA challenges. To perform this evaluation over existing QA datasets, we first extract a large pool of candidate answers from the dataset. Models are then evaluated on their ability to correctly retrieve and rank answers to individual questions using two metrics, \emph{mean reciprocal rank} (MRR) and \emph{recall at N} (R@N)\@. In Eq~(\ref{MRReq}), $Q$ is the set of questions, and \emph{rank}$_i$ is the rank of the first correct answer for the $i$th question. In Eq~(\ref{RNeq}), $A^*_i$ is the set of correct answers for the $i$th question, and $A_i$ is a scored list of answers provided by the model, from which the top N are extracted.

\begin{equation}
\label{MRReq}
\textrm{MRR} = \frac{1}{|Q|} \sum_{i=1}^{|Q|} \frac{1}{\textit{rank}_i}
\end{equation}

\begin{equation}
\label{RNeq}
\textrm{R@N} = \frac{1}{|Q|} \sum_{i=1}^{|Q|} \frac{|\textrm{max}_N(A_i) \cap A^*_i|}{|A^*_i|}
\end{equation}

We explore using the ReQA evaluation on both SQuAD~1.1 and Natural Questions. However, the technique is general and can be applied to other datasets as well.

\subsection{ReQA~SQuAD}
\label{sec:squad}

SQuAD~1.1 is a reading comprehension challenge that consists of over 100,000 questions composed to be answerable by text from Wikipedia articles.
The data is organized into paragraphs, where each paragraph has multiple associated questions.
Each question can have one or more answers in its paragraph.\footnote{Typically, multiple answers come from the same sentence. For example, the question ``Where did Super Bowl 50 take place?''\ is associated with three answers found within the sentence ``The game was played on February 7, 2016, at Levi's Stadium in the San Francisco Bay Area at Santa Clara, California.'' The answer spans are: [Santa Clara, California], [Levi's Stadium] and [Levi's Stadium in the San Francisco Bay Area at Santa Clara, California.].}

We choose SQuAD~1.1 for our initial ReQA evaluation because it is a widely studied dataset, and covers many question types.\footnote{SQuAD 2.0 adds questions that have no answer in the paragraph. While these questions are useful for testing machine reading over fixed passages, their value in a large-scale retrieval evaluation is less clear. Specifically, we can't be sure that such questions aren't answered by another sentence in the larger corpus.}
To turn SQuAD into a retrieval task, we first split each paragraph into sentences using a custom sentence-breaking tool included in our public release.
For the SQuAD~1.1 train set, splitting 18,896 paragraphs produces 91,707 sentences.
Next, we construct an ``answer index'' containing each sentence as a candidate answer.
The model being evaluated computes an answer embedding for each answer (using any encoding strategy), given only the sentence and its surrounding paragraph as input.
Crucially, this computation must be done independently of any specific question.
The answer index construction process is described more formally in Algorithm~\ref{alg:index}.

\begin{algorithm}
  \caption{Constructing the answer index
    \label{alg:index}}
  \begin{algorithmic}[1]
    \Require{$c$ is a representation of a dataset in SQuAD format\footnotemark; $S$ is a function that accepts a string of text, $s$, and returns a sequence of sentences, $[s_0, s_1, \cdots, s_n]$; $E_a$ is the embedding function, which takes answer text, $a$, into points in $\mathbb{R}^n$.}
    \Ensure{A list of $\langle$sentence, encoding$\rangle$ tuples.}
    \Statex
    \Function{EncodeIndex}{$c$, $S$, $E_a$}
      \Let{I}{\textrm{new list}}
      \For{$x \textrm{ in } c\textrm{.data}$} \Comment{for every passage}
        \For{$p \textrm{ in } x\textrm{.paragraphs}$}
            \For{$s$ \textrm{in} $S(p.\textrm{context})$}
              \Let{$s_e$}{$E_a(s, p.\textrm{context})$}
              \State append $\langle s, s_e\rangle$ to I
            \EndFor
        \EndFor
      \EndFor
      \State \Return{I}
    \EndFunction
  \end{algorithmic}
\end{algorithm}
\footnotetext{The SQuAD JSON format consists of a top-level list of \textit{data} elements that represent Wikipedia articles, each containing a list of \textit{paragraphs}. Every paragraph defines a \textit{context}, which is its text, and a corresponding list of questions and answers. For clarity, the algorithm definition uses the same names (lines 3-6).}

Similarly, we embed each question using the model's question encoder, with the restriction that only the question text be used.
For the SQuAD~1.1 train set, this gives around 88,000 questions.

After all questions and answers are encoded, we compute a ``relevance score'' for each question-answer pair by taking the dot-product of the question and answer embeddings, as shown in Algorithm \ref{alg:scoring}.
These scores can be used to rank all (around 92,000) candidate answers for each question, and compute standard ranking metrics such as mean reciprocal rank (MRR) and recall (R@k).\footnote{Rarely, the same question is asked in different contexts. For example, the question ``How tall is Mount Olympus?\@'' appears twice in SQuAD, with answers on the pages for both Greece and Cyprus. In this case, we consider both answers correct for the purposes of our evaluation metrics.}

\begin{algorithm}
  \caption{Scoring questions and answers
    \label{alg:scoring}}
  \begin{algorithmic}[1]
    \Require{$Q_{[q \times n]}$ is a matrix of question embeddings in $\mathbb{R}^n$, arranged so that the $i$-th row, $Q[i]$, corresponds to the embedding of $q_i$; $A_{[a \times n]}$ is a matrix of answer embeddings, also in $\mathbb{R}^n$, derived from the answer index, $I$, and arranged so that the $i$-th row, $A[i]$, corresponds to the embedding of $a_i$.}
    \Ensure{$R_{[q \times a]}$ a matrix of ranking data that can be used to compute metrics such as MRR and R@k. It is arranged so that $i$-th row is a vector of dot-product scores for $q_i$, that is, $[q_i \cdot a_0, q_i \cdot a_1, \cdots, q_i \cdot a_a]$}
    \Statex
    \Function{Score}{$Q$, $A$}
      \Let{$S_{[q \times a]}$}{$QA^T$} \Comment{compute dot-products}
      \Let{$R_{[q \times a]}$}{\textrm{new matrix}}
      \For{$i \leftarrow 1 \textrm{ to } q$}
        \Let{$R[i]$}{$\textrm{rankdata\footnotemark}(S[i])$}
      \EndFor
      \State \Return{$R$}
    \EndFunction
  \end{algorithmic}
\end{algorithm}
\footnotetext{This function assigns ranks to data, in this case assigning 1 to the largest dot-product, 2 to the second-largest dot-product, and so forth. For more details, see \href{https://docs.scipy.org/doc/scipy/reference/generated/scipy.stats.rankdata.html}{scipy.stats.rankdata}.}


\subsection{ReQA~NQ}

Natural Questions (NQ) consists of over 320,000 examples, where each example contains a question and an entire Wikipedia article.
The questions are real questions issued by multiple users to the Google search engine, for which a Wikipedia page appeared in the top five search results.
The examples are annotated by humans as to whether the returned article contains an answer to the question, and if so where.
For roughly 36\% of examples, the article is found to contain a ``short answer'': a span of text (or rarely multiple spans) that directly answers the question.

Our procedure for converting NQ into a ReQA task is similar to that described for SQuAD above.
We restrict to questions with a single-span short answer, contained within an HTML $<$P$>$ (paragraph) block, as opposed to answers within a list or table.
When applied to the NQ training set, this filtering produces around 74,000 questions.
As with SQuAD, we consider the enclosing paragraph as context (available for the model in building an answer embedding), and split the paragraph into sentences.
The target answer is the sentence containing the short answer span.
Each sentence in the paragraph is added to the answer index as a separate answer candidate, resulting in around 240,000 candidates overall.\footnote{Since NQ includes the entire Wikipedia article, we could consider adding all sentences from \emph{all} paragraphs as candidate answers. However even restricting to sentences from paragraphs containing short answers already produced a large index and challenged existing models, so we opted not to increase the search space further.}

As with ReQA~SQuAD, \textit{we advocate excluding all of Wikipedia from model training materials}.
Models satisfying this restriction give us more confidence that they can be extended to perform answer retrieval in new domains.


\subsection{Dataset Statistics}

The number of questions and candidate answers in the ReQA~SQuAD and ReQA~NQ datasets is shown in Table \ref{tab:dataset_counts}.
While the number of questions is similar, ReQA~SQuAD has around 2.6x fewer candidate answer sentences, making it an easier task overall.
This difference is due to the fact that SQuAD itself was constructed to have many different questions answered by the same Wikipedia paragraphs.

\begin{table}[!htb]
\centering
    \begin{tabular}{l r r}
    \toprule
    & {\bf SQuAD} & {\bf NQ} \\
    \midrule
    Questions & 87,599 & 74,097 \\
    Candidate Sentences & 91,707 & 239,013 \\
    Candidate Paragraphs & 18,896 & 58,699 \\
    \bottomrule
    \end{tabular}
\caption{The number of questions and candidates in the constructed datasets ReQA~SQuAD and ReQA~NQ.}
\label{tab:dataset_counts}
\end{table}

Table \ref{tab:dataset_stats} lists the average number of tokens in question and sentence-level answer text, as well as the ``query coverage'', which is the percentage of tokens in the question that also appear in the answer.
The token coverage for ReQA~SQuAD is much larger than for ReQA~NQ, indicating more lexical overlap between the question and answer.
This is likely due to the original SQuAD construction process whereby writers ``back-wrote'' questions to be answerable by the given documents.
By comparison, NQ questions are naturally occurring anonymized, aggregated search queries, where users had no access to the answering document ahead of time.

Table \ref{tab:datatset_types} shows the distribution of question types for each dataset.
Nearly half~(47.7\%) of ReQA~SQuAD questions are \textit{what} questions, with the next most frequent being \textit{who}~(9.6\%) and \textit{how}~(9.3\%).
ReQA~NQ is more balanced across question types, with the leading types being \textit{who}~(32.6\%), \textit{when}~(20.3\%) and \textit{what}~(15.3\%).

We note that neither dataset contains many \textit{why} questions.
Performing well on this type of question may require additional reasoning ability, so it would be interesting to explore \textit{why} questions further through more targeted ReQA datasets.

\begin{table}[!htb]
\centering
    \begin{tabular}{l r r}
    \toprule
    & {\bf SQuAD} & {\bf NQ} \\
    \midrule
    \multicolumn{3}{l}{\textit{Average Length (tokens)}} \vspace{2pt} \\
    \quad Question & 10.1 & 9.1 \\
    \quad Answer & 24.0 & 22.9 \\
    \midrule
    \multicolumn{3}{l}{\textit{Query Coverage (\%)}} \vspace{2pt} \\
    \quad Mean & 31.7 & 24.3 \\
    \quad Standard Deviation & 18.9 & 16.9 \\
    \bottomrule
    \end{tabular}
\caption{Token-level statistics of the constructed datasets. \textbf{Average Length} is the average number of tokens in the question and sentence-level answer text. \textbf{Query Coverage} is the percentage of tokens in the question that also appear in the sentence-level answer.}
\label{tab:dataset_stats}
\end{table}

\begin{table}[!htb]
\centering
    \begin{tabular}{l r r }
    \toprule
    {\bf Question Type}\hspace{1.1cm} & {\bf SQuAD} & {\bf NQ} \\
    \midrule
    \textit{what}    & 47.7 & 15.3 \\
    \textit{who}     &  9.6 & 32.6 \\
    \textit{how}     &  9.3 & 5.0 \\
    \textit{when}    &  6.2 & 20.3 \\
    \textit{which}   &  5.5 & 2.0 \\
    \textit{where}   &  3.8 & 13.1 \\
    \textit{why}     &  1.4 & 0.6 \\
    \midrule
    \textit{other}  & 16.5 & 11.1 \\
    \bottomrule
    \end{tabular}
\caption{The distribution of question types in ReQA~SQuAD and ReQA~NQ\@. A question is assigned to a question type if it starts with the question type word. Note, types \textit{what} and \textit{which} include questions where a preposition (e.g.~\textit{at, by, in, on, with}) appears before the \textit{wh-} word.}
\label{tab:datatset_types}
\end{table}

\subsection{Discussion}
\label{sec:discussion}

A defining feature of the SQuAD dataset is that the questions are ``back-written'', with advance knowledge of the target answer and its surrounding context.
One concern when adapting this data for a ReQA task is that questions may become ambiguous or underspecified when removed from the context of a specific document and paragraph.
For example, SQuAD~1.1 contains the question ``What instrument did he mostly compose for?''.
This question makes sense in the original context of the Wikipedia article on Frédéric Chopin, but is underspecified when asked in isolation, and could reasonably have other answers.
One possible resolution would be to include the context title as part of the question context.
However this is unrealistic from the point of view of end systems where the user doesn't have a specific document in mind.

This concern can be avoided by switching from ``back-written'' datasets to ``web-search based'' datasets.
These include MS MARCO \citep{msmarco}, TriviaQA \citep{JoshiTriviaQA2017} and Natural Questions \citep{natural-questions}.
For these sets, questions are taken from natural sources, and a search engine is used in the process of constructing QA pairs.

However, there is an important caveat to mention when using web-search data to build ReQA tasks.
In these datasets, the answers are derived from web documents retrieved by a search engine, where the question is used as the search query.
This introduces a bias toward answers that are already retrievable through traditional search methods.
By comparison, answers in SQuAD~1.1 may be found in ``off-topic'' documents, and it is valuable for an evaluation to measure the ability to retrieve such answers.
Since both types of datasets (back-written and web-search based) have their advantages, we believe there is value in evaluating on ReQA tasks of both types.


\section{Models and Results}
\label{sec:results}

In this section we evaluate neural models and classic information retrieval techniques on the ReQA~SQuAD and ReQA~NQ benchmark tasks.


\subsection{Neural Baselines}

\begin{figure}[!htb]
  \centering
  \includegraphics[width=0.35\textwidth]{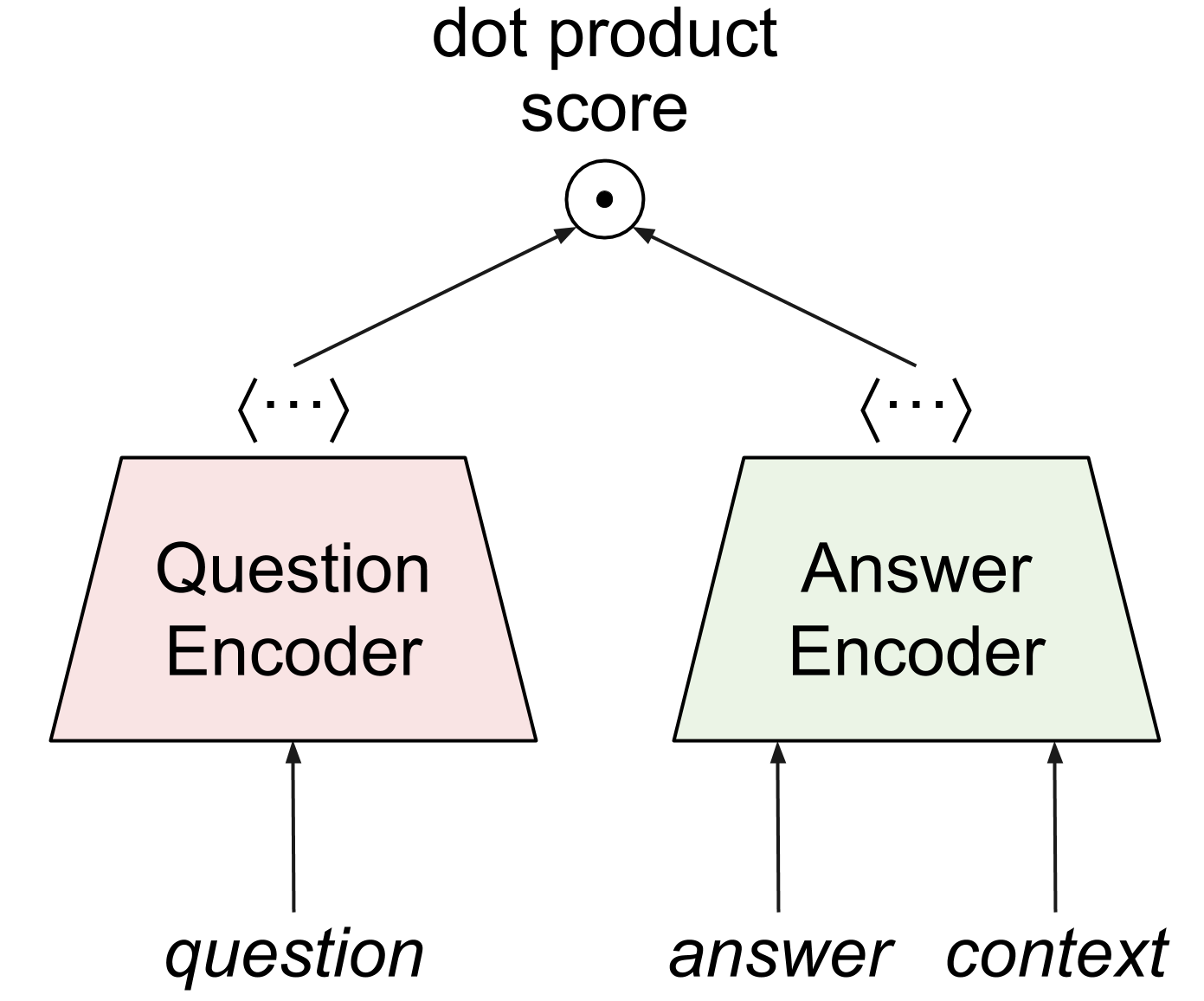}
  \caption{
    A schematic dual encoder for question-answer retrieval.
  }
  \label{fig:dual_encoder}
\end{figure}

Dual encoder models are learned functions that collocate queries and results in a shared embedding space.
This architecture has shown strong performance on sentence-level retrieval tasks, including conversational response retrieval~\citep{DBLP:journals/corr/HendersonASSLGK17,yang-etal-2018-learning}, translation pair retrieval~\citep{guo-etal-2018-effective,DBLP:journals/corr/abs-1902-08564} and similar text retrieval~\citep{DBLP:journals/corr/abs-1811-08008}. A dual encoder for use with ReQA has the schematic shape illustrated in Figure \ref{fig:dual_encoder}.

As our primary neural baseline, we take the recently released universal sentence encoder QA (USE-QA) model from \citet{useqa}\footnote{\url{https://tfhub.dev/google/universal-sentence-encoder-multilingual-qa/1}}.
This is a multilingual QA retrieval model that co-trains a question-answer dual encoder along with secondary tasks of translation ranking and natural language inference.
The model uses sub-word tokenization, with a 128k ``sentencepiece'' vocabulary \citep{sentencepiece}.
Question and answer text are encoded independently using a 6-layer transformer encoder \citep{vaswani2017}, and then reduced to a fixed-length vector through average pooling.
The final encoding dimensionality is 512.
The training corpus contains over a billion question-answer pairs from popular online forums and QA websites like Reddit and StackOverflow.

As a second neural baseline, we include an internal QA model (QA\textsubscript{Lite}) designed for use on mobile devices.
Like USE-QA, this model is trained over online forum data, and uses a transformer-based text encoder.
The core differences are reduction in width and depth of model layers, reduction of sub-word vocabulary size, and a decrease in the output embedding size from 512 dimensions to only 100.

Finally, we include the text embedding system InferSent, which, although not explicitly designed for question answering tasks, nevertheless produces strong results on a wide range of semantic tasks without requiring additional fine-tuning~\citep{conneau-EtAl:2017:EMNLP2017}. Note, however, that at 4096 dimensions, its embeddings are significantly larger than the other baselines presented. Other systems in this class include Skip-thought~\citep{Kiros:2015:SV:2969442.2969607}, ELMo~\citep{Peters:2018}, and the Universal Sentence Encoder\footnote{The non-QA versions of the Universal Sentence Encoder produce general semantic embeddings of text.}.

Table \ref{tab:sentence} presents the ReQA results for our baseline models.
As expected, the larger USE-QA model outperforms the smaller QA\textsubscript{Lite} model.
The recall@1 score of 0.439 on ReQA~SQuAD indicates that USE-QA is able to retrieve the correct answer from a pool of 91,707 candidates roughly 44\% of the time.
The ReQA~NQ scores are lower, likely due to both the larger pool of candidate answers, as well as the lower degree of lexical overlap between questions and answers.

Table \ref{tab:resources} illustrates the tradeoff between model accuracy and resource usage.

\begin{table}[!htb]
\centering
    \begin{tabular}{l c c c c}
    \toprule
    {\bf Model} & {\bf MRR} & {\bf R@1} & {\bf R@5} & {\bf R@10} \\
    \midrule
    \multicolumn{5}{l}{\textit{ReQA~SQuAD}} \vspace{2pt} \\
    \quad USE-QA & 0.539 & 0.439 & 0.656 & 0.727 \\
    \quad QA\textsubscript{Lite} & 0.412 & 0.325 & 0.507 & 0.576 \\
    \quad InferSent & 0.317 & 0.240 & 0.402 & 0.468 \\
    \midrule
    \multicolumn{5}{l}{\textit{ReQA~NQ}} \vspace{2pt} \\
    \quad USE-QA & 0.234 & 0.147 & 0.317 & 0.391 \\
    \quad QA\textsubscript{Lite} & 0.172 & 0.103 & 0.233 & 0.297 \\
    \quad InferSent & 0.080 & 0.043 & 0.109 & 0.145 \\
    \bottomrule
    \end{tabular}
\caption{Mean reciprocal rank (MRR) and recall@K performance of neural baselines on ReQA~SQuAD and ReQA~NQ.}
\label{tab:sentence}
\end{table}

\begin{table}[!htb]
\centering
    \begin{tabular}{l r r r}
    \toprule
    {\bf Model} & {\bf Size} & {\bf Latency\footnotemark} & {\bf Memory} \\
    & (MB) & (ms) & (MB) \\ 
    \midrule
    USE-QA & 392.9 & 17.3 & 71.8 \\
    QA\textsubscript{Lite} & 2.6 & 10.2 & 3.6 \\
    \bottomrule
    \end{tabular}
\caption{Time and space tradeoffs of different models. Latency was measured on an Intel Xeon CPU E5-1650 v3 @ 3.50GHz, which has 6 cores and 12 threads.}
\label{tab:resources}
\end{table}

\footnotetext{This is the latency for encoding a single piece of text. However, by batching the encoding requests, it's possible to significantly reduce the amortized encoding time. In practice,
batch sizes of 200 provide an amortized speedup of up to 5x.}

\subsection{BM25 Baseline}

While neural retrieval systems are gaining popularity, TF-IDF based methods remain the dominant method for document retrieval, with the BM25 family of ranking functions providing a strong baseline~\citep{Robertson:2009}.
Unlike the neural models described above that can directly retrieve content at the sentence level, such methods generally consist of two stages: document retrieval, followed by sentence highlighting~\citep{DBLP:journals/ftir/MitraC18}.
Previous work in open domain question answering has shown that BM25 is a difficult baseline to beat when questions were written with advance knowledge of the answer~\citep{lee2019latent}.

To obtain our baseline using traditional IR methods, we constructed a paragraph-level retrieval task which allows a direct comparison between the neural systems in Table \ref{tab:sentence} and BM25.\footnote{We opted not to evaluate BM25 on sentence-level retrieval as earlier work has shown that traditional term-based document retrieval technologies are unsuccessful when applied to sentence-level retrieval~\cite{Allan:2003:RND:860435.860493}.}
We evaluate BM25 by measuring its ability to recall the paragraph containing the answer to the question.\footnote{Our experiments make use of the implementation at \url{https://github.com/nhirakawa/BM25} with default hyperparameter settings.}
To get a paragraph retrieval score for our neural baselines, we run sentence retrieval as before, and use the retrieved sentence to select the enclosing paragraph.
As shown in Table \ref{tab:paragraph}, the USE-QA neural baseline outperforms BM25 on paragraph retrieval.

\begin{table}[!htb]
\centering
    \begin{tabular}{l c c c c}
    \toprule
    {\bf Model} & {\bf MRR} & {\bf R@1} & {\bf R@5} & {\bf R@10} \\
    \midrule
    \multicolumn{5}{l}{\textit{ReQA~SQuAD}} \vspace{2pt} \\    
    \quad USE-QA & \textbf{0.634} & \textbf{0.533} & \textbf{0.756} & \textbf{0.823} \\
    \quad QA\textsubscript{Lite} & 0.503 & 0.407 & 0.613 & 0.689 \\
    \quad InferSent & 0.369 & 0.279 & 0.469 & 0.548 \\
    \quad BM25\footnotemark & 0.602 & 0.517 & 0.702 & 0.755 \\
    \midrule
    \multicolumn{5}{l}{\textit{ReQA~NQ}} \vspace{2pt} \\
    \quad USE-QA & \textbf{0.366} & \textbf{0.247} & \textbf{0.486} & \textbf{0.578} \\
    \quad QA\textsubscript{Lite} & 0.274 & 0.177 & 0.366 & 0.450 \\
    \quad InferSent & 0.145 & 0.082 & 0.199 & 0.258 \\
    \quad BM25 & 0.103 & 0.066 & 0.140 & 0.175 \\    
    \bottomrule
    \end{tabular}
\caption{Performance of various models on paragraph-level retrieval.}
\label{tab:paragraph}
\end{table}

\footnotetext{BM25 statistics were computed over the first 10,000 questions of each dataset, due to slow scoring speed.}


\section{Related Work}
\label{sec:related}

Open domain question answering is the problem of answering a question from a large collection of documents~\citep{trec}.
Successful systems usually follow a two-step approach to answer a given question: first retrieve relevant articles or blocks, and then scan the returned text to identify the answer using a reading comprehension model~\citep{Jurafsky:2009:SLP:1214993,kratzwald-feuerriegel-2018-adaptive,yang-etal-2019-end,lee2019latent}.
While the reading comprehension step has been widely studied with many existing datasets~\citep{rajpurkar-etal-2016-squad,msmarco,searchQA,natural-questions}, machine reading at scale is still a challenging task for the community.

\citet{chen-etal-2017-reading} recently proposed DrQA, treating Wikipedia as a knowledge base over which to answer factoid questions from SQuAD~\citep{rajpurkar-etal-2016-squad}, CuratedTREC~\citep{curatedTREC} and other sources.
The task measures how well a system can successfully extract the answer span given a question, but it still relies on a document retrieval step.
The ReQA eval differs from DrQA task by skipping the intermediate step and retrieving the answer sentence directly.

There is also a growing interest in answer selection at scale. 
\citet{surdeanu-etal-2008-learning} constructs a dataset with 142,627 question-answer pairs from Yahoo!~Answers, with the goal of retrieving the right answer from all answers given a question.
However, the dataset is limited to ``how to'' questions, which simplifies the problem by restricting it to a specific domain.
Additionally the underlying data is not as broadly accessible as SQuAD and other more recent QA datasets, due to more restrictive terms of use.

WikiQA~\citep{yang-etal-2015-wikiqa} is another task involving large-scale sentence-level answer selection.
The candidate sentences are, however, limited to a small set of documents returned by Bing search, and is smaller than the scale of our ReQA tasks. WikiQA consists of 3,047 questions and 29,258 candidate answers, while ReQA~SQuAD and ReQA~NQ each contain over 20x that number of questions and over 3x that number of candidates (see Table~\ref{tab:dataset_counts}).
Moreover, as discussed in Section~\ref{sec:discussion}, restricting the domain of answers to top search engine results limits the evaluation's applicability for testing end-to-end retrieval.

\citet{DBLP:journals/corr/abs-1810-10176} made use of SQuAD for a retrieval task at the paragraph level. We extend this work by investigating sentence level retrieval and by providing strong sentence-level and paragraph-level baselines over a replicable construction of a retrieval evaluation set from the SQuAD data. Further, while \citet{DBLP:journals/corr/abs-1810-10176} trained their model on data drawn from SQuAD, we would like to highlight that our own strong baselines do not make use of any training data from SQuAD\@. We advocate for future work to attempt a similar approach of using sources of model training and evaluation data that are distinct as possible in order to provide a better picture of how well models generally perform a task.

Finally, \citet{seo-etal-2018-phrase} construct a phrase-indexed question answering challenge that is similar to ReQA in requiring the question and the answer be encoded separately of one another. However, while ReQA focuses on sentence-based retrieval, their benchmark retrieves phrases, allowing for a direct $F_1$ and exact-match evaluation on SQuAD\@. \citet{seo-etal-2019-real} demonstrate an implementation of a phrase-indexed question answering system using a combination of dense (neural) and sparse (term-frequency based) indices.

We believe that ReQA can help guide development of such systems by providing a point of evaluation between SQuAD, whose passages are too small to test retrieval performance, and SQuAD-Open \citep{chen-etal-2017-reading}, which operates at a realistic scale but is expensive and slow to evaluate. In practice, our evaluation runs completely in memory and finishes within two hours on a developer workstation, making it easy to integrate directly into the training process, where it can, for instance, trigger early stopping.

\section{Conclusion}

In this paper, we introduce Retrieval Question-Answering (ReQA) as a new benchmark for evaluating end-to-end answer retrieval models.
The task assesses how well models are able to retrieve relevant sentence-level answers to queries from a large corpus. 
We describe a general method for converting reading comprehension QA tasks into cross-document answer retrieval tasks.
Using SQuAD and Natural Questions as examples, we construct the ReQA~SQuAD and ReQA~NQ tasks, and evaluate several models on sentence- and paragraph-level answer retrieval.
We find that a freely available neural baseline, USE-QA, outperforms a strong information retrieval baseline, BM25, on paragraph retrieval, suggesting that end-to-end answer retrieval can offer improvements over pipelined systems that first retrieve documents and then select answers within.
We release our code for both evaluation and conversion of the datasets into ReQA tasks.

\section*{Acknowledgments}

We thank our teammates from Descartes and other Google groups for their feedback and suggestions. We would like to recognize Javier Snaider, Igor Krivokon and Ray Kurzweil. Special thanks goes to Jonni Kanerva for developing the tailored sentence-breaking algorithm.

\bibliography{emnlp-ijcnlp-2019}
\bibliographystyle{acl_natbib}

\appendix

\end{document}